\documentclass[sigconf, nonacm]{acmart}

\usepackage{algorithm}
\usepackage{algorithmic}

\AtBeginDocument{%
  }

\setcopyright{acmlicensed}
\copyrightyear{2025}
\acmYear{2025}
\acmDOI{XXXXXXX.XXXXXXX}

\acmConference[XXX '25]{XXXX}{XXX, 2025}{XXXX}
\acmISBN{978-1-4503-XXXX-X/18/06}

\begin{document}
\title{Causal Predictive Optimization and Generation for Business AI}


\author{Liyang Zhao}
\authornote{These authors contributed equally.}
\affiliation{%
\institution{LinkedIn Corporation}
\city{Sunnyvale}
\country{USA}}
\email{liyzhao@linkedin.com}

\author{Olurotimi Seton}
\authornotemark[1]
\affiliation{%
\institution{LinkedIn Corporation}
\city{Sunnyvale}
\country{USA}}
\email{tseton@linkedin.com}

\author{Himadeep Reddy Reddivari}
\authornotemark[1]
\authornotemark[2]
\affiliation{%
\city{Seattle}
\country{USA}}
\email{hreddy1203@gmail.com}

\author{Suvendu Jena}
\affiliation{%
\institution{LinkedIn Corporation}
\city{Sunnyvale}
\country{USA}}
\email{sjena@linkedin.com}

\author{Shadow Zhao}
\affiliation{%
\city{Redwood City}
\country{USA}}
\authornote{Work done while at LinkedIn.}
\email{shadow19900519@gmail.com}

\author{Rachit Kumar}
\affiliation{%
\institution{LinkedIn Corporation}
\city{Sunnyvale}
\country{USA}}
\email{rackumar@linkedin.com}

\author{Changshuai Wei}
\authornote{Corresponding author}
\affiliation{%
\institution{LinkedIn Corporation}
\city{Seatttle}
\country{USA}}
\email{chawei@linkedin.com}

\renewcommand{\shortauthors}{Zhao et al.}

\begin{abstract}
The sales process involves converting leads or opportunities into customers and selling additional products to existing clients. Optimizing this process is therefore key to the success of any B2B business. In this work, we introduce a principled approach to sales optimization and business AI, \textit{Causal Predictive Optimization and Generation}, which comprises three layers: (1) a prediction layer using causal ML; (2) an optimization layer with constraint optimization and contextual bandits; and (3) a serving layer featuring Generative AI and feedback loop. We detail the implementation and deployment of this system at LinkedIn, and share learnings and insights broadly applicable to the field.
\end{abstract}




\maketitle
\sloppy 

\section{Introduction}
Sales functions play a critical role in the success and revenue growth of any Business-to-Business (B2B) or Software-as-a-Service (SaaS) company. Each member of the sales team manages anywhere from a handful to hundreds of accounts, and their primary objective is to convert leads and upsell existing customers. Sales representatives spend the majority of their time identifying the right account or contact to engage, and deciding what to discuss. Determining the optimal approach for each account is highly time-consuming. Across the B2B and SaaS sectors, sales teams commonly encounter these challenges.
\begin{itemize}
    \item Inefficient allocation of resources: Sales reps often spend excessive time on low-impact accounts while overlooking high-value opportunities;
    \item Complex decision trade-offs: Balancing targets for revenue growth, customer engagement, and capacity constraints without clear guidance slows down decision-making;
    \item Low trust and adaptability in AI-driven recommendations: Opaque, rigid systems impede adoption, and static models rarely incorporate real-time feedback, undermining long-term effectiveness.
\end{itemize}
An efficient business AI engine can greatly improve sales-team productivity and is key to optimizing the sales process and driving business success. In this paper, we propose \textit{Casual Predictive Optimization and Generation (CPOG)} as an end-to-end framework solution.

\subsection{Related Work}

Machine learning (ML) models play a crucial role in account prioritization by guiding sales teams toward high-potential prospects based on behavioral patterns and engagement signals. These models also improve sales forecasting by leveraging both structured data \cite{rohaan2022using} and textual data \cite{elalem2023machine}, and customer segmentation through clustering techniques facilitates better sales planning by identifying key audience groups \cite{gautam2022customer}. B2B customer churn prediction \cite{de2021uplift} and e-commerce customer conversion models \cite{gubela2019conversion} further demonstrate the value of predictive analytics in sales. However, most existing prediction models do not exploit causal ML methods \cite{rubin1978bayesian, kennedy2023towards, van2006statistical, gutierrez2017causal} for estimating the individual heterogeneous effects of potential sales actions, despite promising industry applications in other areas \cite{shi2024ads, tang2023counterfactual}.

Separately, optimization frameworks have been developed to turn predictive scores into actionable recommendations—examples include neural optimization with adaptive heuristics (NOAH) for intelligent marketing systems \cite{wei2024neural} and regression and chaotic pattern search (RCPS) for lead generation \cite{gaddam2024data}. Contextual bandit approaches have also shown value for personalized recommendations \cite{ban2024neural}. Yet these methods were not designed for end-to-end sales optimization and no existing optimization framework seamlessly integrates into the sales process.

Meanwhile, advances in explainable AI (XAI) and generative AI (GAI) are beginning to transform how businesses consume ML outputs. Model-agnostic, rule-based explanations improve interpretability \cite{bohanec2017explaining, rajapaksha2022limref}, and GAI techniques promise more natural interactions with sales engines.

Despite these advances, there is currently no comprehensive solution that unifies causal prediction, constrained optimization, explainability, and generative serving at enterprise scale for B2B and SaaS sales. The only two frameworks that (remotely) resemble our approach are NOAH \cite{wei2024neural} and RCPS \cite{gaddam2024data}. NOAH targets intelligent marketing rather than sales optimization and does not include explainable AI or generative serving. RCPS addresses only lead generation as a sub-process of sales optimization, and does not incorporate causal ML, multi-objective optimization, XAI, or GAI, nor has it been demonstrated at industry scale.




\subsection{Our Contribution}

To our knowledge, CPOG is the first end-to-end framework for sales optimization in business AI. Moreover, it is the first sales optimization engine deployed in an industry setting that:
\begin{itemize}
    \item Utilizes causal machine learning to measure the incremental impact of sales actions in the prediction layer, avoiding inefficient use of sales resources;
    \item Features a dedicated optimization layer to trade off multiple objectives and constraints in the sales process, and recommend actions that balance exploration and exploitation;
    \item Serves recommendation with explainability, GAI components and feedback-loop, forming a cohesive human-in-the-loop systems that works seamlessly with sales teams.
\end{itemize}

\section{System Overview}
CPOG tackles sales-ecosystem challenges with a structured, AI-driven architecture comprising three complementary layers: the \textbf{Prediction Layer}, which applies causal machine-learning models to estimate both short- and long-term impacts of potential sales-rep interventions, enabling more efficient resource allocation and clearer ROI insights; the \textbf{Optimization Layer}, which integrates business constraints and prioritization logic via mixed-integer programming and contextual bandit algorithms to recommend actions that balance multiple objectives (e.g., revenue growth, engagement) while managing exploration–exploitation trade-offs; and the \textbf{Serving Layer}, which delivers these recommendations with human-interpretable explanations, embedded Generative AI components for natural interactions, and a continuous feedback loop that refines the system over time.

Together, these layers form a cohesive human-in-the-loop system that streamlines account prioritization, personalizes outreach, and maximizes revenue impact. In the following sections, we provide a detailed discussion of each layer and its components.



\begin{figure}[h]
  \centering
  \includegraphics[width=\linewidth]{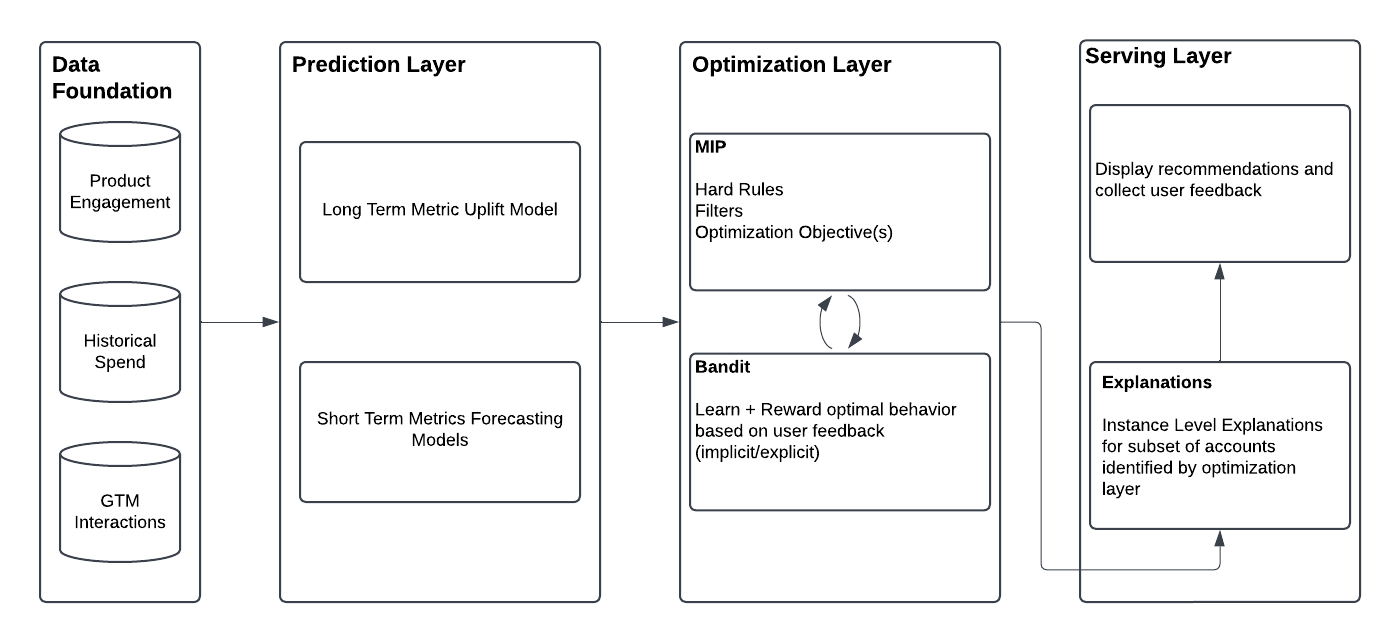}
  \caption{Overall CPOG Architecture}
  \Description{Overall CPOG Architecture}
  \label{CPOG_arch}
\end{figure}



\section{Prediction Layer}

The prediction layer predicts a group of performance metrics which sales functions can influence through sales actions. Denoting the feature for customers as $x$, feature of possible sales actions as $a$ and the performance metrics as $y$, the prediction layer builds a functional mapping:
\begin{equation*}
    y = f(x, a)
\end{equation*}
We consider two types of metrics: (i) monetization metrics with longer feedback periods like revenue, customer loyalty and customer LTV, (ii) customer engagement metrics like product utilization and product adoption with shorter feedback periods. The goal of the prediction layer is to estimate the impact of sales actions on monetization metrics and predict near term customer engagement metrics. The functional mapping $f(.)$ can be learned from historical data on customer outcomes. 

\subsection{Monetization Uplift Models}

Monetization uplift models refer to the set of casual ML models predicting the incremental impact of sale action on monetization metrics. Since sales functions have limited human resources, it is crucial that these resources are put on highest incremental impact. Conventional approach of directly predicting the monetization metrics and recommending accounts with highest ranks can result in approaching accounts that will convert without sales outreach and thus waste of sales resources. \\
Let $y$ denote the desired outcome of certain monetization metric, $a$ the sales outreach action, and $x^U$  the pre-treatment variables e.g product utilization, revenue before the treatment etc.
We predict $y$ using ML models, i.e. $y_a=f(x,a)$. The nature of the ML models used depends on the quantity of data: a simple regression model will suffice for smaller datasets, tree-based models for medium sized data and neural networks for large and complex datasets. The uplift from sales outreach when $a\in \{0,1\}$ can estimated as follows:
\begin{equation*}
    y^U = y_1 - y_0.
\end{equation*}
Besides above "two" learner (T-learner) formulation, we can also adopt other Meta-learners\cite{athey2016recursive}\cite{kunzel2019metalearners} methods such as S-learner, X-learner or DR-learner. Further information are in  
Appendix~\ref{apx:uplift}.
 
\subsection{Customer Engagement Forecast Models}
The second objective of the prediction layer is estimating customer engagement metrics, $E$, like product utilization and product adoption. 
\begin{equation*}
\begin{split}
    y^E = f(x^E)
\end{split}
\end{equation*}
Where $x^E$ are customer engagement features like prior customer engagement metrics, customer firmographics etc.
If a product has low data volume e.g. a newly launched product, simple regression models will suffice. As more data and signals become available, more advanced regression models like gradient boosted trees or neural networks might be better suited.

\section{Optimization Layer}
The optimization layer transforms predictive scores into actionable sales recommendations using a dual strategy that combines constrained optimization with contextual bandit methods. The \textit{constrained optimization} component balances key business objectives, such as maximizing revenue uplift and engagement, against predefined operational constraints, systematically determining the optimal account–representative assignments to align with strategic goals. Simultaneously, the \textit{contextual bandit} component adapts daily action recommendations in real time, focusing on immediate priorities like boosting engagement, preventing churn, and driving upsells. Together, these techniques enable the system to respond dynamically to evolving market conditions and sales-rep interactions.

\subsection{Constrained Optimization}
\subsubsection{Formulating the Constrained Optimization}
Let $\mathcal{C} = \{c_1, c_2, \dots, c_N\}$ represent the set of customer accounts and $\mathcal{R} = \{r_1, r_2, \dots, r_M\}$ the set of sales representatives. Define the binary decision variable $a_{i,j} \in \{0, 1\}$ to indicate whether account $c_i$ is assigned to representative $r_j$. The objective is to maximize a weighted combination of monetization uplift ($U_i$) and engagement improvement ($E_i$).

\begin{equation}
\max_{a_{i,j}} \sum_{i=1}^N \sum_{j=1}^M \Big[ w(d_i) \cdot y^{U}_i + (1 - w(d_i)) \cdot y^{E}_i \Big] \cdot a_{i,j},
\end{equation}

where,
\begin{itemize}
  \item $y^{U}_i$: Monetization uplift for account $a_i$ (normalized to range 0 -100).
  \item $y^{E}_i$: A function that represents Engagement difference for account $a_i$. e.g.
  \begin{equation}
  y^{E}_i = \max\left(|\Delta y^{E1}_i|, |\Delta y^{E2}_i|, \ldots, |\Delta y^{Ek}_i|\right),
\end{equation}

where \( \Delta y^{Ej}_i \) denotes the change in the \(j\)-th engagement metric for the \(i\)-th account, and \( k \) is the total number of engagement metrics considered.
(normalized to range 0 -100).
  \item $d_i$: Days until RTCD (Renwal Target Close Date) for account $a_i$.\footnote{RTCD (Renewal Target Close Date) refers to the date by which a renewal decision for an account is targeted to be concluded. It is a critical date for sales reps, guiding prioritization and resource allocation to ensure timely engagement and renewal success.}
  \item $w(d_i)$: A non-linear Weighting function for $d_i$. In this example, it is defined as:
  \begin{equation}
  w(d_i) = \frac{1}{1 + e^{-k(d_i - d_0)}},
  \end{equation}
  where $k$ controls the steepness of the curve and $d_0$ centers it.
\end{itemize}

\subsubsection{Constraints}

The optimization adheres to the following constraints:
\begin{enumerate}
    \item \textbf{Capacity Constraint:} Each sales representative $r_j$ manages between $n_{\text{min}}$ and $n_{\text{max}}$ accounts:
    \begin{equation}
    n_{\text{min}} \leq \sum_{i=1}^N a_{i,j} \leq n_{\text{max}}, \quad \forall i \in \mathcal{C}, \forall j \in \mathcal{R}.
    \end{equation}
    
    \item \textbf{Assignment Constraint:} Each account is assigned to exactly one representative:
    \begin{equation}
    \sum_{j=1}^M a_{i,j} = 1, \quad \forall i \in \mathcal{C}, \forall j \in \mathcal{R}.
    \end{equation}
    
    \item \textbf{Eligibility Filters:} Accounts must satisfy predefined thresholds:
    \begin{itemize}
        \item $y^{U}_i > T_U$: Monetization uplift exceeds threshold $T_U$.
        \item $y^{E}_i > T_E$: Engagement improvement exceeds threshold $T_E$.
\item For any account \( c_i \) that has been assigned (\( a_{ij} = 1 \)) within the last 14 days, ensure it is not assigned again:
\[
a_{ij} = 0, \quad \forall j \in \mathcal{R}, \text{ if } \sum_{t=T-14}^{T-1} z_{i,t} > 0, \quad \forall i \in \mathcal{C}
\]
Where
\[
z_{i,t} = 
\begin{cases} 
1 & \text{if } a_{ij} = 1 , \exists j\in\mathcal{R} \text{ at time t} \\
0 & \text{otherwise}
\end{cases}
\]
\end{itemize}  
    \item \textbf{RTCD Priority:} Monetization uplift is prioritized for accounts with $d_i$ close to zero, while engagement is emphasized for large $d_i$, which is contolled by  as $w(d_i)$ in (2).
\end{enumerate}

\subsubsection{Recommendation Rules}

The original problem is formulated as a Mixed Integer Programming (MIP) problem. We "relax" the problem to Linear Programming by changing binary deterministic decision variable $a_{ij}\in\{0,1\}$ to a probability of taking action, i.e., $0 \leq a_{ij} \leq 1$, for two reason:
\begin{enumerate}
\item \textbf{Ranking}: Besides matching of accounts $c$ to reps $r$, we also have needs to form a ranking of account to inform reps the priority on the set of matched account. 
\item \textbf{Computation}: Directly solving MIP can be computational expensive and are more likely to have convergence problem. Relaxed LP has advantage on computation efficiency and pipeline stability.
\end{enumerate}

We recommend the accounts to reps with following 2 steps:
\begin{itemize}
    \item \textbf{Account Matching and Ranking:} Matching is done by rounding $a_{ij}$, i.e., accounts \( c_i \) is matched to sales reps \( r_j \) if $\lfloor a_{ij} + 0.5\rfloor = 1$. Among the set of matched accounts, we rank them by $a_{ij}$. See example results in Table~\ref{tab:optimization_outputs_details}. 
    \item \textbf{Action Recommendation:} Heuristic algorithms are utilized to decide initial action recommendation on the matched account, based on values of \( y^{U}_i \) and \( y^{E}_i \). (see Algorithm ~\ref{algo:action} and example result in Table~\ref{tab:optimization_outputs_actions}). This serves as a cold start for contextual bandit ranking in section ~\ref{sec:cb}. 
\end{itemize}


\subsubsection{Implementation Notes}
 
\begin{itemize}
    \item \textbf{Weight Calibration:} The weighting function $w(d_i)  = \frac{1}{1 + e^{-k(d_i - d_0)}} $ balances monetization and engagement based on days to RTCD. We fix \( d_0 = X \times 30 \) days (domain knowledge) and tune sharpness \( k \) via backtesting to align recommendations with renewal outcomes. This offers interpretability through a fixed anchor and empirical flexibility.
    \item \textbf{Solver Strategy:} LP can be solved via SCIP \cite{ortools, BolusaniEtal2024OO} in offline manner for typical sale optimization scale. \textit{Dualip} \cite{pmlr-v119-basu20a} or ML-augmented solvers~\cite{bertsimas2022online} can be used for large scale or low-latency use cases. 
\end{itemize}


\begin{algorithm}
\caption{Decision Making Process for Action Recommendation}
\begin{algorithmic}
\label{algo:action}
\FOR{each account $a_i$}
    \STATE Calculate $MonetizationValue_i = w(d_i) \cdot y^{U}_i$
    \STATE Calculate $EngagementValue_i = (1 - w(d_i)) \cdot y^{E}_i$

    \IF{$MonetizationValue_i \leq EngagementValue_i$}
        \IF{$\min(|\Delta y^{E1}_i|, \dots, |\Delta y^{Ek}_i|) \geq \max(|\Delta y^{E1}_i|, \dots, |\Delta y^{Ek}_i|)$}
            \STATE Recommend ``Boost Engagement''
        \ELSE
            \STATE Recommend ``Promote Upsell''
        \ENDIF
    \ELSE
        \IF{$y^{U}_i > 0$}
            \STATE Recommend ``Promote Upsell''
        \ELSE
            \STATE Recommend ``Prevent Churn''
        \ENDIF
    \ENDIF
\ENDFOR
\end{algorithmic}
\end{algorithm}

\begin{table}[h]
  \caption{Optimization Layer Outputs - Account Details}
  \label{tab:optimization_outputs_details}
  \centering
  \begin{tabular}{cccc}
    \toprule
    \textbf{Account ID} & \textbf{Rep ID} & \textbf{gRank}\footnotemark[1] & \textbf{rRank}\footnotemark[2] \\
    \midrule
    101 & R1 & 1 & 1 \\
    102 & R1 & 2 & 2 \\
    103 & R2 & 3 & 1 \\
    \bottomrule
  \end{tabular}
\end{table}
\footnotetext[1]{\textbf{gRank} \text{:account ranking across all reps based on $a_{ij}$}}
\footnotetext[2]{\textbf{rRank} \text{:account ranking within the assigned rep based on $a_{ij}$}}

\begin{table}[h]
  \caption{Optimization Layer Outputs - Actions and Metrics}
  \label{tab:optimization_outputs_actions}
  \begin{tabular}{cccc}
    \toprule
    \textbf{Account ID} & \textbf{Action} & \textbf{\(U_i\)} & \textbf{\(E_i\)} \\
    \midrule
    101 & Promote Upsell & 5000 & 15 \\
    102 & Boost Engagement & 2000 & 30 \\
    103 & Prevent Churn & -1000 & 5 \\
    \bottomrule
  \end{tabular}
\end{table}

\subsection{Contextual Bandit}
\label{sec:cb}

After constrained optimization, the recommendations are processed by a contextual bandit layer, which decide final action recommendation among \texttt{Boost\_Engagement}, \texttt{Prevent\_Churn} and \texttt{Promote\_Upsell}. The contextual bandit iteratively refines its policy, dynamically balancing exploration (to improve understanding of less tried recommendations) and exploitation (to serve the most effective recommendations).

\subsubsection{Model Set-up}
Each sales representative is assigned a set of accounts determined by the constraint optimization step. The contextual bandit selects an action $a_t\in \mathcal{A}$ to maximize cumulative reward over time:
\[
\max_{a_t\in \mathcal{A}_t} \mathbb{E} \left[ \sum_{t=1}^T y(a_t, x_t) \right],
\]
where, $\mathcal{A}_t = \{a^B, a^C, a^U \}$, \( a^B \) index \texttt{Boost\_Engagement}, \( a^C \) \texttt{Prevent\_Churn} and \( a^U \) \texttt{Promote\_Upsell}; \( x_t \) represents contextual features at time \( t \), and \( y(a_t, x_t) \) denotes the reward from user feedback, with reward defined as 
\[
y =
\begin{cases}
    +1 & \text{if user feedback = \text{DEEP\_LINK\_CLICKED},} \\
    -1 & \text{if user feedback = \text{NOTIFICATION\_DISMISSED},} \\
    0 & \text{if user feedback = \text{NO\_CLICK}.}
\end{cases}
\]
\subsubsection{Policy Optimization}

Policy optimization is achieved by employing Neural Bandit \cite{dai2022sample, zhou2020neural}. Each action's reward distribution is modeled as $
y_a = h(x_t, a_t) + \epsilon$, where $h(x_t, a_t)$ is a unkown functional mapping from feature to the expected reward, $\epsilon$ is  a sub-Gaussian noise.

In Neural Bandit, we approximate the unknown function $h(\cdot)$ with a neural network $f_{\theta}(\cdot)$. The algorithm iterate over these steps:
\begin{enumerate}
    \item Compute contextual feature representation $x_t$.
    \item Predict reward for each action $a\in \mathcal{A}$ using the neural network:
    \[
    \hat{y}_a= f_{\theta}(x_t, a) + \delta.
    \]
    \item Select the action that maximizes the predicted reward:
    \[
    a_t = \arg \max_{a \in \mathcal{A}} \hat{y}_a.
    \]
    \item Observe user feedback and update network parameters $\theta$ via backpropagation.
\end{enumerate}
Here, value of $\delta$ depends on a key variance quantity $\sigma$,
$$\sigma = \sqrt{\nabla f(\cdot)^\top H^{-1} \nabla f(\cdot)}$$
where \( \nabla f(\cdot) \) is the gradient of the neural network output with respect to input features, \( H^{-1} \) is the inverse Hessian matrix used to capture model uncertainty. 

In Thompson Sampling, we sample $\delta$ from Gaussian distribution, $\delta \sim \mathcal{N}(0, \beta^2\sigma^2)$, 
and in Upper Confidence Bound, we calculate 
$\delta=\gamma\sigma$, 
where both $\beta$ and $\gamma$ are tuning parameter controlling the balancing of exploration and exploitation.

Through these steps, the contextual bandit ensures that engine evolve to serve the most relevant recommendation to the user effectively while optimizing long-term engagement.

\subsubsection{Contextual Feature Representation}

The contextual bandit layer incorporates a rich set of features to enhance decision-making. The contextual feature vector is defined as:

\[
x_t = [x^A_t, x^S_t, x^R_t],
\]

where:
\begin{itemize}
    \item \( x^A_t \) (Account Features): Includes attributes such as account size, industry, engagement level.
    \item \( x^S_t \) (Sales Representative Features): Encapsulates experience, historical success rate, and past interactions.
\end{itemize}

These features are extracted using:
\[
x_{\text{numerical}} = 
\begin{bmatrix}
\text{TimeSinceLastAlert} \\
\text{PreviousAlertCount}
\end{bmatrix},
\]
with one-hot encoded categorical representations:
\[
x_{\text{categorical}} = \text{Encode}(\text{Feedback Type, Alert Category, User Metadata}).
\]

\section{Serving Layer}
The Serving Layer in the System serves the recommendation to sales functions and further transforms predictive scores and features into actionable insights for sales representatives. By employing a structured and interpretable methodology, it ensures recommendations are clear, precise, and aligned with business objectives. Our framework has been designed to support template based explanations and template-free explanations powered by GAI.

\subsection{Template based explanations}

\subsubsection{Formulating the Explanation Layer}

The Explanation Layer is built on three primary components: feature mapping, threshold definitions, and templates. It translates model outputs into human-readable insights through:
\begin{enumerate}
\item Mapping feature names to human-understandable expressions.
\item Defining thresholds for key features and model combinations.
\item Generating recommendations based on these mappings and thresholds.
\end{enumerate}
Examples for feature mapping and threshold definition can be found in Appendix Table~\ref{tab:feature_mapping} and Table~\ref{tab:basic_info_thresholds}.
\subsubsection{Templates for Recommendations}

Templates are designed to produce tailored recommendations for each account. These recommendations are categorized into key alert types and include structured explanations for feature values that pass thresholds. Examples include:

\begin{itemize}
    \item \textbf{Low Engagement:}
    RTCD = $d_1$. We recommend reaching out to the client to understand the low engagement in <product>. The current usage is $y$ and we are predicting a drop to $\Delta y$ over the next month.

    \item \textbf{Upsell Flag:}
    RTCD = $d_2$. We recommend exploring add-on opportunities with this customer as we predict a near-term upsell opportunity worth $\Delta y$.

    \item \textbf{Churn Flag:}
    RTCD = $d_3$. We recommend connecting with customers to assess churn risks.

\end{itemize}

\subsection{Template-free explanations with GAI}

\begin{figure}[h]
  \centering
  \includegraphics[width=0.5\textwidth]{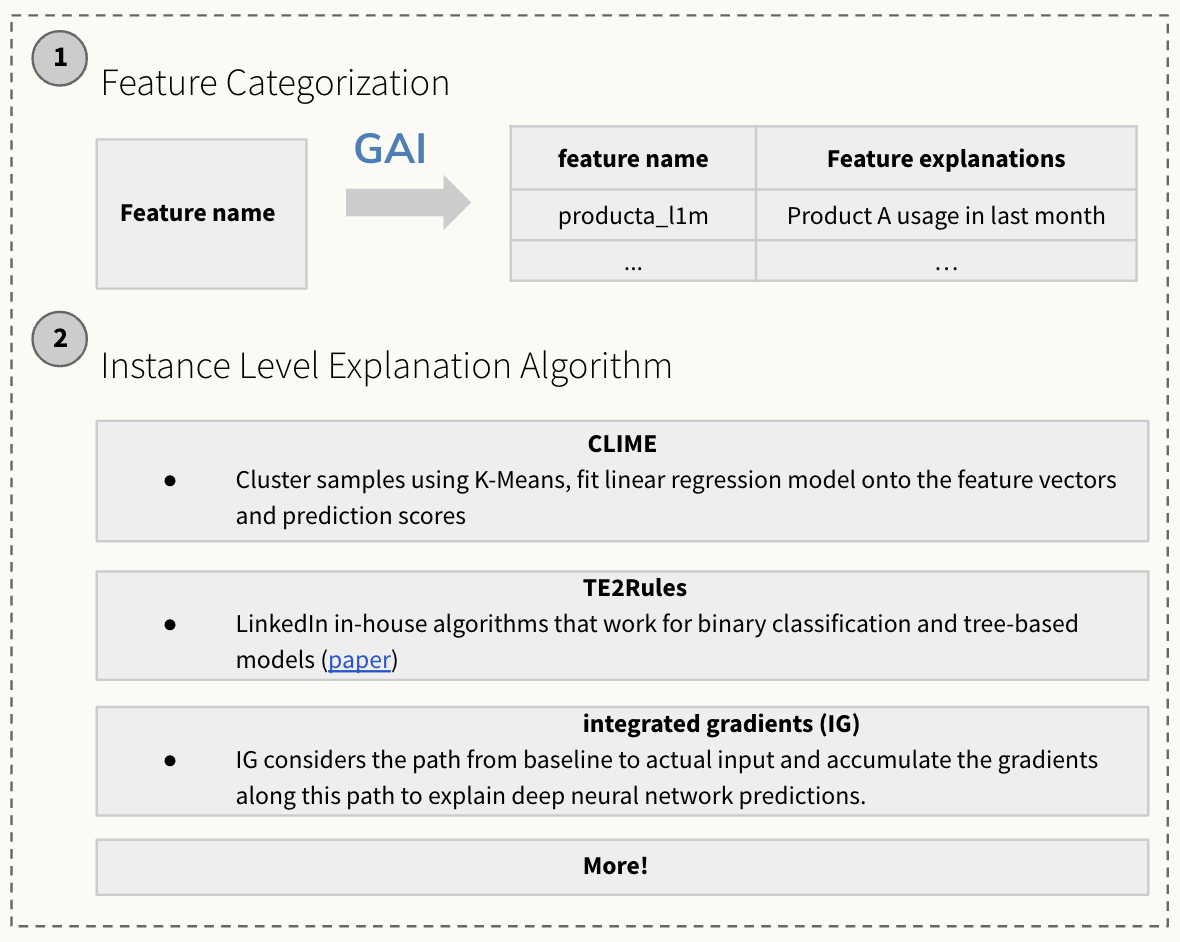}
  \captionsetup{width=0.45\textwidth}
  \caption{GAI Based Instance Level Explanation Generation Process}
  \label{fig:gai}
\end{figure}

To enhance the explanation layers and streamline the on-boarding process, we leverage LLM to generate feature explanations for each feature name, and integrate Generative AI (GAI) with sales productivity tools to automate narrative generation and feature categorization.  Multiple instance-level explanation algorithms (e.g., CLIME, TE2Rules, Integrated Gradients) are incorporated for calculating the importance of instance-level features (Figure ~\ref{fig:gai}). Using an iterative approach, we specify the necessary steps to complete a task and provide a sample output format for GAI to learn. By integrating feature names and explanations along with instance-level feature importance scores into the LLM, we generate detailed instance-level explanations. This includes insights into what changes in each feature might indicate, offering a comprehensive understanding of the data's impact on model recommendations (Figure ~\ref{fig:gai_example}). Detais are provided in Appendix~\ref{appendix:gai-process}.

\begin{figure}[h]  
  \centering
  \includegraphics[width=0.5\textwidth]{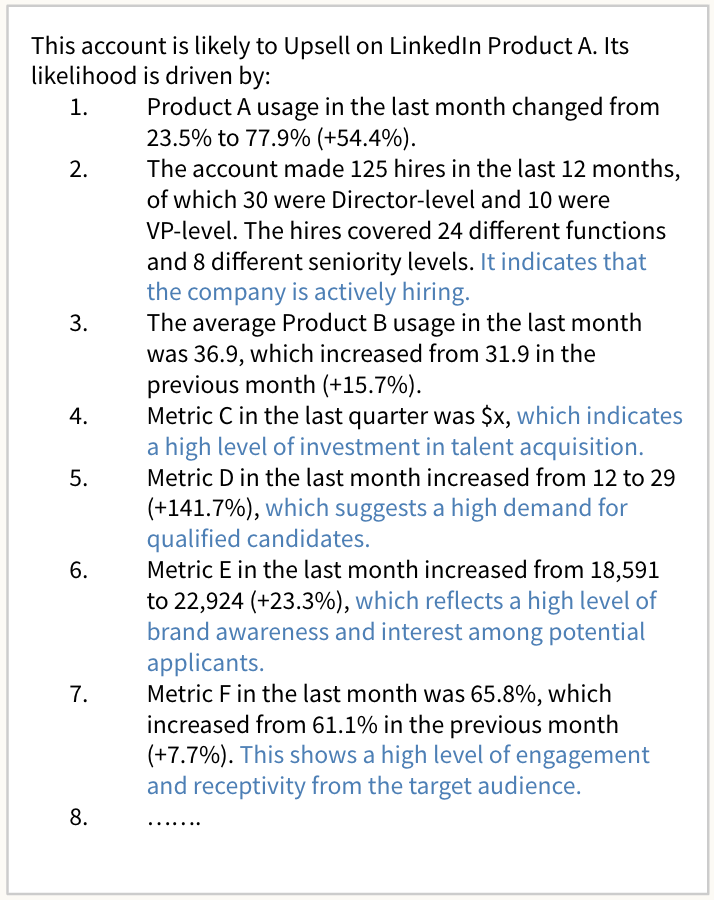}  
  \captionsetup{width=0.45\textwidth}  
  \caption{GAI Based Instance Level Explanation Generation Process}
  \label{fig:gai_example}
\end{figure}%

\begin{figure}[h]
  \centering
  \includegraphics[width=0.5\textwidth]{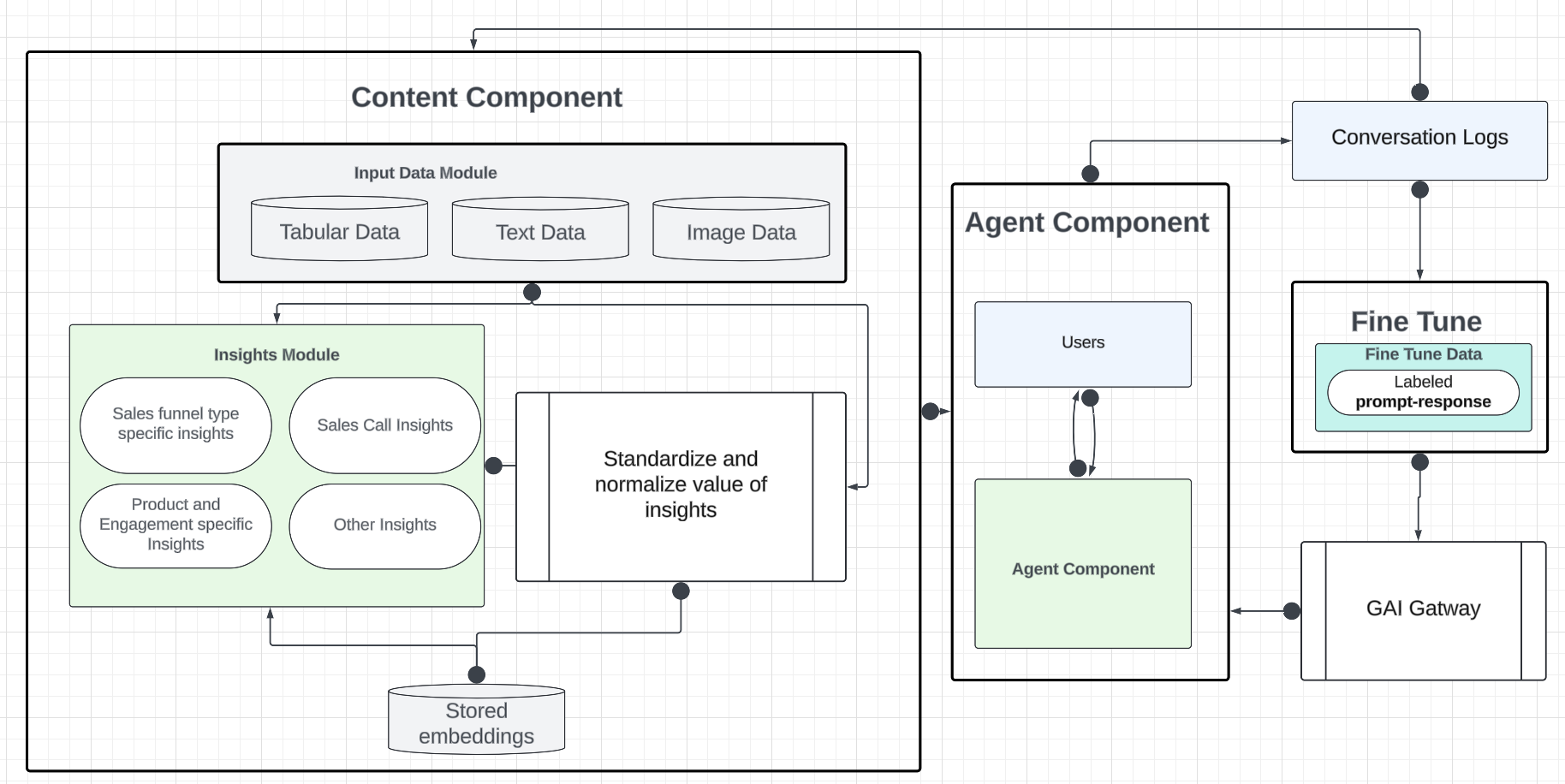}
  \captionsetup{width=0.45\textwidth} 
  \caption{GAI Based Instance Level Explanation Generation Process}
  \label{fig:chat_bot}
\end{figure}

We also integrate multiple components to generate insights and provide conversational experience for user in agentic AI framework (Figure ~\ref{fig:chat_bot}). The Content Component processes input data, including tabular, text, and image data, through an Input Data Module. This data is then used in the Insights Module to generate sales funnel-specific, sales call, product and engagement insights, and other relevant insights, which are standardized and normalized. The Agent Component interacts with users, utilizing these insights and generating conversation logs. These logs, including labeled prompt-response pairs, are used in the Fine-Tune module to improve the agentic AI system. 

\section{Experiments and Results}

\subsection{Offline Analysis}
We evaluate the performance of each layers of CPOG in offline Analysis. We provide high level summary here. Details can be found in Appendix~\ref{apx:offline}. 

\subsubsection{Offline Evaluation for Prediction Models}
\begin{figure}[h]
  \centering
  \includegraphics[width=0.5\textwidth]{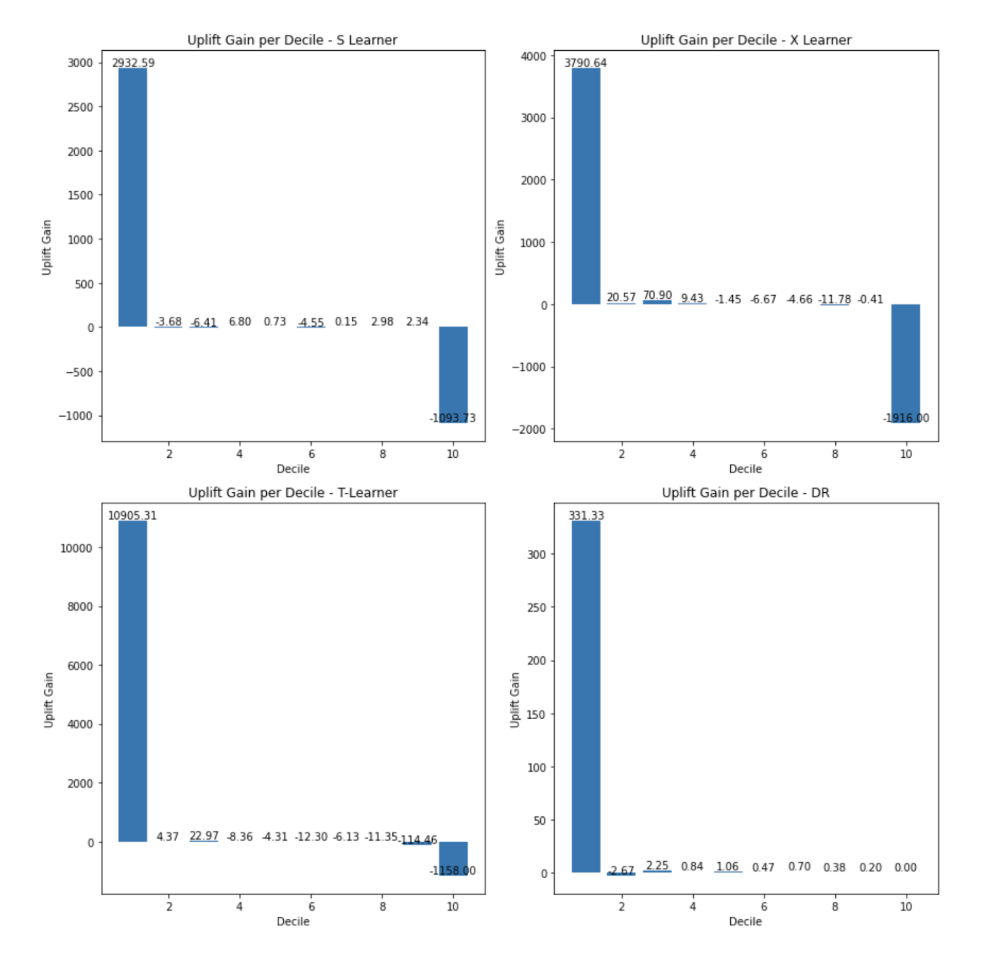}
  \captionsetup{width=0.45\textwidth}  
  \caption{Uplift Deciles for Baseline Models}
  \label{fig:uplift}
\end{figure}

We evaluated 4 uplift estimators: S-learner, T-learner, X-learner, and DR-Learner, by ranking customers into uplift deciles\cite{gutierrez2017causal} and examining uplift curves (Figure \ref{fig:uplift}), while also considering prediction stability and score interpretability. Although the S- and X-learners produced sensible ITE distributions, we ultimately selected the T-learner for its consistently stable decile bins and ITE estimates across multiple runs. 

For customer engagement, we modeled monthly product adoption ($pa$) and utilization ($pu$) and measured performance via Mean Absolute Error (MAE) and Root Mean Squared Error (RMSE); the results for models $f_{pu}$ and $f_{pa}$ are summarized in Table \ref{tab:engagement_model_perf}.

\begin{table}[h]
  \caption{Forecasting models prediction accuracy metrics}
  \label{tab:engagement_model_perf}
  \begin{tabular}{cccl}
    \toprule
    Model & Score Range & MAE & RMSE\\
    \midrule
    $f_{pu}$ & 0- 100& 7.581 & 10.775\\
    $f_{pa}$ & 0 - 1& 0.203 & 0.291\\
  \bottomrule
\end{tabular}
\end{table}

\subsubsection{Ablation analysis for Constrained Optimization}
We conducted an ablation analysis of the optimization layer by systematically removing or modifying key components—such as the weighting function, capacity constraints, and recommendation rules. Using six months of backtesting data, we compared the performance of the Full Model (with all components intact) to simplified models across key metrics, including upsell precision, churn precision, churn recovery precision, low engagement precision, total bookings rank, and constraint feasibility. The Full Model consistently outperformed the ablated models, demonstrating the importance of each component. (Details in Appendix ~\ref{sec:abalation})

\subsubsection{Validation for Contextual Bandit}
We validated the contextual bandit layer via offline simulations that assessed its recommendation quality and learning behavior. First, we analyzed the feedback distribution which illustrates the bandit’s ability to prioritize recommendations that results in high-reward signals. We then examined feedback trends over time, revealing the transition from exploration to exploitation as the model increasingly favors actions yielding greater rewards. Finally, we compared cumulative rewards for Neural Thompson Sampling and Neural UCB to evaluate policy convergence and exploration–exploitation balance; both methods achieved similar overall returns, but we selected Thompson Sampling for production due to its lower per-iteration computational cost and resulting low serving latency. (Details in Appendix~\ref{apx:bandit})

\subsection{Online Experiment}

We launched an online test within LinkedIn, targeting sales reps that focused on SME customers in North America. Sales reps received personalized recommendations which are delivered in near real-time through CRM applications. Our solution received high user satisfaction among sales reps (details in Appendix ~\ref{sec:explanatin_result}) and delivered significant business impact (+59.96\% lift on primary metrics) with minimal disruption to sales workflows and quota achievement.

\subsubsection{Challenges on randomized A/B Test}
While randomized A/B test is the best practice for impact estimation, several challenges must be considered in our scenario: (i) \textit{Opportunity Costs,} A partial rollout could limit solution availability, resulting in missed opportunities despite its potential to alleviate sales representatives' pain points with minimal disruption; (ii) \textit{Business Urgency}, Achieving revenue goals amidst macroeconomic pressures necessitates leveraging the solution’s timely insights promptly; (iii) \textit{Selection Bias:} Randomization in treatment group selection may be infeasible, introducing potential selection bias; (iv) \textit{Quota Equity}, Disparities in quotas arising from the test must be mitigated through appropriate quota adjustments during interim sales planning cycles;
(vi) \textit{Statistical Power}, Testing with a small group may lack sufficient power to produce statistically significant results.

\subsubsection{Observational Study}
Given the challenges of conducting a randomized A/B test, we implemented an observational causal study to estimate the business impact. The treatment and control groups were defined as follows:
\begin{itemize}
    \item \textbf{Treatment Group:} Sales representatives in North America who had access to the recommendations
    \item \textbf{Control Group:} Sales representatives outside North America without access to the recommendations
\end{itemize}

\noindent The units in our treatment group are <rep, account> pairs because recommendations are generated for only a subset of accounts within each sales reps' books.
We focus exclusively on \textit{relevant booking opportunities}\footnote{Opportunities originating from accounts with valid recommendations, generated within a specified time period, and followed up by sales outreach} which makes our treatment period time-dependent. To account for irregular and sparse timing of treatments across multiple periods, we utilized Difference-in-Difference (DiD) \cite{ashenfelter1984using} methodology for our causal analysis. Formally, the DiD estimator is defined as,
$$
\widehat{\tau}_{\text{DiD}}
= \bigl(\bar Y_{\text{treat},1} - \bar Y_{\text{treat},0}\bigr)
- \bigl(\bar Y_{\text{ctrl},1} - \bar Y_{\text{ctrl},0}\bigr)
$$
which measures the treatment effect under the parallel‐trends assumption. We can calculate the relative treatment effect by,
$$
\text{RTE}
= \frac{\widehat{\tau}_{\text{DiD}}}{\bar Y_{\text{ctrl},1} - \bar Y_{\text{ctrl},0}},
$$
Details on assumption validation are in Appendix~\ref{apx:did}.

\subsubsection{Experiment Setup}
Our online experiments consists of two time periods: (1) A/A - 6 months before intervention  and (2) observational (non-randomized) A/B - 6 months after experiment started. 
Our control group consists of candidate accounts outside North America and were matched to the treatment group using Coarsened Exact Matching (CEM) \cite{iacus2012causal}. 
We used a proxy metric, Net Ratio, as the primary response variable for our test period and define Net Ratio as follows:
\begin{equation}
    \text{Net Ratio (NR)} = \frac{\text{renewal AND add-on bookings}}{\text{renewal target amount}}
    \label{eq:net_ratio}
\end{equation}

\subsubsection{Results}

The causal inference analysis demonstrated a significant improvement in the primary metric, Net Ratio (NR), when compared to the legacy system. As defined in Equation~\ref{eq:net_ratio}, Net Ratio captures account-level monetization effectiveness.



\begin{table}[h]
  \caption{Difference-in-Difference Results on Net Ratio (NR)}
  \label{tab:did}
  \begin{tabular}{ccc}
    \toprule
    Method  & Relative Treatment Effect on NR & p-value\\
    \midrule
    Legacy& N/A & N/A\\
    CPOG&  59.96\% & 0.0178\\
  \bottomrule
\end{tabular}
\end{table}

\section{Conclusion}
In this paper, we introduce the Causal Predictive Optimization and Generation (CPOG) framework, a novel approach to sales optimization and business AI. Through offline analysis, online testing and deployment, we demonstrate CPOG's effectiveness in optimizing sales strategies across multiple metrics.  In particular, we highlight the importance of using uplift models to prioritize sales actions, the benefits of constrained optimization in balancing multiple objectives, and the value of providing sales representatives with clear, explainable and actionable recommendations. We give guidelines on how to apply CPOG framework and clear examples of how we implemented it at LinkedIn, along with results and potential measurement approaches in complex applications where randomized A/B testing is not feasible. The framework and the learnings in deployments are broadly applicable to the field, particular B2B and SaaS businesses.


\bibliographystyle{ACM-Reference-Format}
\bibliography{CPOG}

\appendix
\section{Additional Details on Method}

\subsection{Uplift Modeling}
\label{apx:uplift}
\begin{itemize}
    \item \textbf{S-learner} \cite{kunzel2019metalearners} uses a single estimator that includes all features and treatment indicators without giving the treatment indicator a special role. While this simplified structure makes the S-learner easy to implement, not assigning the indicator a special role means the model may or may not use the indicator feature for modeling.

    \item \textbf{T-learner} \cite{kunzel2019metalearners} or Two-learner is the most common meta-algorithm for estimating heterogeneous treatment effects by building two regression models trained on the control and treatment groups separately. The treatment effect is estimated as the difference between the predictions from the regression models. T-learners can be biased if either regression model is trained on insufficient data
    
    \item \textbf{X-learner} \cite{kunzel2019metalearners} extends the concept of T-learners by using each observation in the training set in an "X"-like shape. This allows for estimating the CATE by regressing the difference of the ITEs on the covariates. X-learner improves on the limitations of other meta-learners - it is more effective for instances where the training data volume for one outcome is greater than the other. The need for large training data is also a limitation of X-learners. 

    \item \textbf{DR Learner} \cite{robins1994estimation} The DR learner estimates treatment effects under the assumption that all confounders are observed by breaking the problem into two predictive tasks: predicting the outcome from treatment and controls, and predicting treatment from controls. It then combines these models in a final stage to estimate heterogeneous treatment effects.
\end{itemize}

\subsection{Explanation tables}

\begin{table}[h]
\caption{Feature Name Mapping}
\label{tab:feature_mapping}
\centering 
\begin{tabular}{p{0.3\linewidth} p{0.6\linewidth}}
\toprule
\textbf{Feature Name} & \textbf{Expression} \\
\midrule
producta\_l1m & Avg. product A usage for last month \\
productb\_l2m & Avg. product B usage across last 2 months \\
\bottomrule
\end{tabular}
\end{table}

\begin{table}[h]
\centering
\caption{Basic Information and Thresholds}
\label{tab:basic_info_thresholds}
\begin{tabular}{lll}
\toprule
\textbf{Feature Name} & \textbf{Model} & \textbf{Threshold} \\
\midrule
producta\_l1m & Treatment Model & >0 \\
producta\_l1m & Control Model & <0 \\
\bottomrule
\end{tabular}
\end{table}


\subsection{GAI-Based Instance-Level Explanation Generation Process}
\label{appendix:gai-process}

This appendix outlines the multi-stage process for generating interpretable, instance-level explanations by combining algorithmic feature importance with semantic grouping.

\subsubsection{Step 1: Feature Grouping using LLM}
Given a list of raw feature names, we use LLM to group them semantically based on suffix patterns and domain expertise:

\begin{itemize}
    \item \textbf{super\_name:} Meaning of features.
    \item \textbf{ultra\_name:} Higher-level category that unifies related features across timeframes.
\end{itemize}

\textbf{Mapping Rules:} Internal shorthand and acronyms are expanded (e.g., metric prefixes) and grouped accordingly. Related engagement metrics (e.g., Metric A, Metric B) are clustered under broader semantic categories.

\textbf{Example Output:}

\begin{table}[H]
\centering
\begin{tabular}{lll}
\toprule
Feature\_name & super\_name & ultra\_name \\
\midrule
MetricA\_l1m & Metric A in the last month & Metric A \\
MetricA\_l2m\_l1m & Metric A in the last month & Metric A \\
MetricA\_l3m & Metric A in the last 3 months & Metric A \\
MetricB\_l1m & Metric B in the last month & Metric B \\
\bottomrule
\end{tabular}
\caption{Abstracted Feature Grouping Example}
\end{table}

\subsubsection{Step 2: Combining Feature Importance with Metadata}

We extract instance-level feature importance from the instance-level feature importance algorithm and join it with grouped feature metadata. This includes:

\begin{enumerate}
    \item Extracting feature importance scores from instance-level feature importance algorithm.
    \item Joining with the semantic feature grouping table (containing \texttt{super\_name} and \texttt{ultra\_name}).
    \item Enriching with feature values per account.
\end{enumerate}

\textbf{Output Schema:}

\begin{table}[H]
\centering
\begin{tabular}{lll}
\toprule
Weight & Feature\_name & Value \\
\midrule
0.072 & MetricC\_l12m & 24.0 \\
\bottomrule
\end{tabular}
\caption{Step 2 output table 1}
\end{table}

\begin{table}[H]
\centering
\begin{tabular}{lll}
\toprule
Super\_name & Ultra\_name & customer\_urn \\
\midrule
Metric C in the last 12 months & Metric C & urn:li:customer:... \\
\bottomrule
\end{tabular}
\caption{Step 2 output table 2}
\end{table}

\subsubsection{Step 3: Explanation Generation using LLM}

We use LLM to generate ranked natural language explanations based on weighted features. The generation logic:

\begin{itemize}
    \item One insight per distinct \texttt{Ultra\_name}, ordered by descending feature weight.
    \item Combine narratives for related metrics (e.g., all hiring-related features grouped into a single insight).
\end{itemize}

\textbf{Example Output:}

\begin{quote}
\small
\texttt{This account is likely a good candidate to Promote Upsell. Its likelihood is driven by: \\
1. In the past 12 months, Metric C increased from 18 to 24 (+33\%).This shows a high level of engagement and receptivity from the target audience. \\
2. Metric A in the last month increased from 78\% to 85\% (+7\%), which reflects a high level of brand awareness and interest among potential applicants.}
\end{quote}
\section{Additional Offline Analysis}
\label{apx:offline}
\subsection{Details of Offline Evaluation for Prediction}
\label{apx:pred}

\subsubsection{Dataset and Features}
Training data for offline evaluation consists of active SME customers, with closed opportunities, during the data collection period. These closed opportunities were either won or disengaged. Our features included customer engagement metrics, historical bookings, product usage, company firmographics, and LinkedIn-only data like hiring trends. 

The predictive models utilized in the prediction layer were trained and evaluated on historical data consisting of product engagement, previous customer transactions, and Go-To-Market (GTM) interaction metrics. GTM interaction metrics include sales outreach, sales offers, and other incentives used to retain or grow customer accounts over time. Other features include company firmographics and LinkedIn's proprietary customer data like hiring trends, among others. The training data are refreshed regularly using LinkedIn proprietary orchestration platform which ensures the prediction layer models are updated and capture concept drift, covariate shift, and label drift over time. The prediction layer also handles data quality concerns, such as noisy data and sparse historical records. Finally, the prediction layer utilizes a robust pipeline for training the model used by the monetization uplift and customer engagement forecast models. Our robust pipeline prevents overfitting the model using cross-validation and utilizes data monitoring tools to detect drift in model performance over time.

\subsubsection{Uplift Model Evaluation}The uplift model estimates the Individual Treatment Effect of sales actions on monetization metrics for each customer. We compared four Conditional Average Treatment Effects (CATE) estimators for the uplift model: S-learner, T-learner, X-learner and DR Learner \cite{kunzel2019metalearners, robins1994estimation, econml2019econml} (Details in Appendix ~\ref{apx:uplift}). 

\noindent Evaluating uplift models can be challenging due to the lack of ground truth for counterfactual outcomes, which makes it difficult to find a loss measure for each observation. 
We use the \textbf{uplift deciles} \cite{gutierrez2017causal} for aggregated measures such as uplift bins or uplift curves(Figure ~\ref{fig:uplift}). Our evaluation also focused on internal business considerations such as prediction stability and score interpretability as secondary metrics.

\noindent While the results from the S-learner and X-learner models mimic the expected ITE distribution, we selected the T-learner for our uplift modeling due to the consistency and stability of the decile bins and ITE distributions over multiple iterations.
\subsubsection{Customer Engagement Models Evaluation}
Customer engagement metrics are continuous values, that measure monthly product adoption and product utilization for each customer. We denote these metrics as $pa$ and $pu$ respectively. We evaluated model accuracy using Mean Absolute Error (MAE) and Root Mean Squared Error (RMSE).
Table \ref{tab:engagement_model_perf} summarizes the prediction accuracy results for our models $f_{pu}$ and $f_{pa}$.

\subsection{Details of Ablation Analysis for Optimization}
\label{sec:abalation}
\subsubsection{Objective}
To evaluate the effectiveness of each primary component in the optimization framework, including feature mappings, constraints, and recommendation rules. The objective is to quantify their individual contributions to monetization outcomes and overall model effectiveness.

\subsubsection{Experimental Setup}
We performed simulations under various configurations by systematically removing or modifying key components of the optimization layer. The performance was evaluated using six months of backtesting data to measure the impact of each component.

\begin{itemize}
    \item \textbf{Full Model (Baseline):} Includes all components—weighted objective function, capacity constraints, RTCD prioritization, and eligibility filters.
    \item \textbf{Ablated Models:}
    \begin{enumerate}
        \item \textbf{Model A (No Weighting Function):} Removed the weighting function \( w(d_i) \), treating monetization uplift and engagement improvement equally across all accounts.
        \item \textbf{Model B (Relaxed Capacity Constraints):} Removed capacity constraints, allowing sales representatives to be assigned any number of accounts.
        \item \textbf{Model C (Simplified Recommendation Rules):} Simplified the recommendation rules to always prioritize monetization uplift.
    \end{enumerate}
\end{itemize}

\subsubsection{Metrics}
\begin{enumerate}
    \item \textbf{Upsell Precision (\( P_{\text{ups}} \)):} The percentage of accounts recommended for upsell that successfully closed as upsell opportunities.
    \item \textbf{Churn Precision (\( P_{\text{ch}} \)):} The percentage of accounts recommended as churn that actually churned without sales outreach.
    \item \textbf{Churn Recovery Precision (\( P_{\text{rec}} \)):} The percentage of accounts recommended as churn that were successfully retained through sales outreach.
    \item \textbf{Low Engagement Precision (\( P_{\text{low}} \)):} The percentage of accounts recommended as low engagement that churned without sales outreach.
    \item \textbf{Total Bookings Rank (\( B_{\text{total}} \)):} The total revenue target achieved (RTA) across all account categories ranked in descending order.
    \item \textbf{Model Feasibility (\% Constraints Met):} The percentage of capacity and assignment constraints satisfied during optimization.
\end{enumerate}

\subsubsection{Results}

Table~\ref{tab:ablation_performance} and~\ref{tab:ablation_ranking} are results from backtesting data.

\begin{table}[t]
\caption{Ablation Analysis Results - Model Performance}
\label{tab:ablation_performance}
\centering
\begin{tabular}{lcccc}
\toprule
\textbf{Model} & \textbf{\(P_{\text{ups}}\)} & \textbf{\(P_{\text{ch}}\)} & \textbf{\(P_{\text{rec}}\)} & \textbf{\(P_{\text{low}}\)} \\
\midrule
Full & 60\% & 55\% & 50\% & 40\% \\
Model A & 56\% & 45\% & 42\% & 30\% \\
Model B & 35\% & 26\% & 37\% & 25\% \\
Model C & 52\% & 43\% & 38\% & 48\% \\
\bottomrule
\end{tabular}
\end{table}

\begin{table}[t]
\caption{Ablation Analysis Results - Ranking and Constraints}
\label{tab:ablation_ranking}
\centering
\begin{tabular}{lcc}
\toprule
\textbf{Model} & \textbf{\(B_{\text{total}}\) Rank} & \textbf{\% Constraints Met} \\
\midrule
Full Model             & 2 & 100\% \\
Model A (No Weighting) & 3 & 100\% \\
Model B (No Constraints) & 1 & 216\% \\
Model C (Simplified Rules) & 4 & 100\% \\
\bottomrule
\end{tabular}
\end{table}

The ablation analysis highlights the critical contributions of each component in the optimization layer to its overall effectiveness. The Full Model's superior performance demonstrates the necessity of incorporating all key features:

\begin{itemize}
    \item \textbf{Weighting Function:} Essential for dynamically adjusting priorities between monetization uplift and engagement improvement based on account proximity to RTCD.
    \item \textbf{Capacity Constraints:} Crucial for ensuring practical and equitable assignment of accounts while maintaining feasibility across operational workflows.
    \item \textbf{Nuanced Recommendation Rules:} Integral to balancing diverse objectives, such as upsell, churn recovery, and engagement improvement, to achieve holistic optimization.
\end{itemize}

While simplified models (e.g., Model B and Model C) may yield partial gains in specific metrics, they compromise overall balance and feasibility. Therefore, the Full Model provides the most robust solution for optimizing sales strategies, ensuring both short-term and long-term business objectives are met effectively.

\subsection{Details of Validation Analysis for Contextual Bandit}
\label{apx:bandit}
\begin{figure}[h]
        \centering\includegraphics[width=0.5\textwidth]{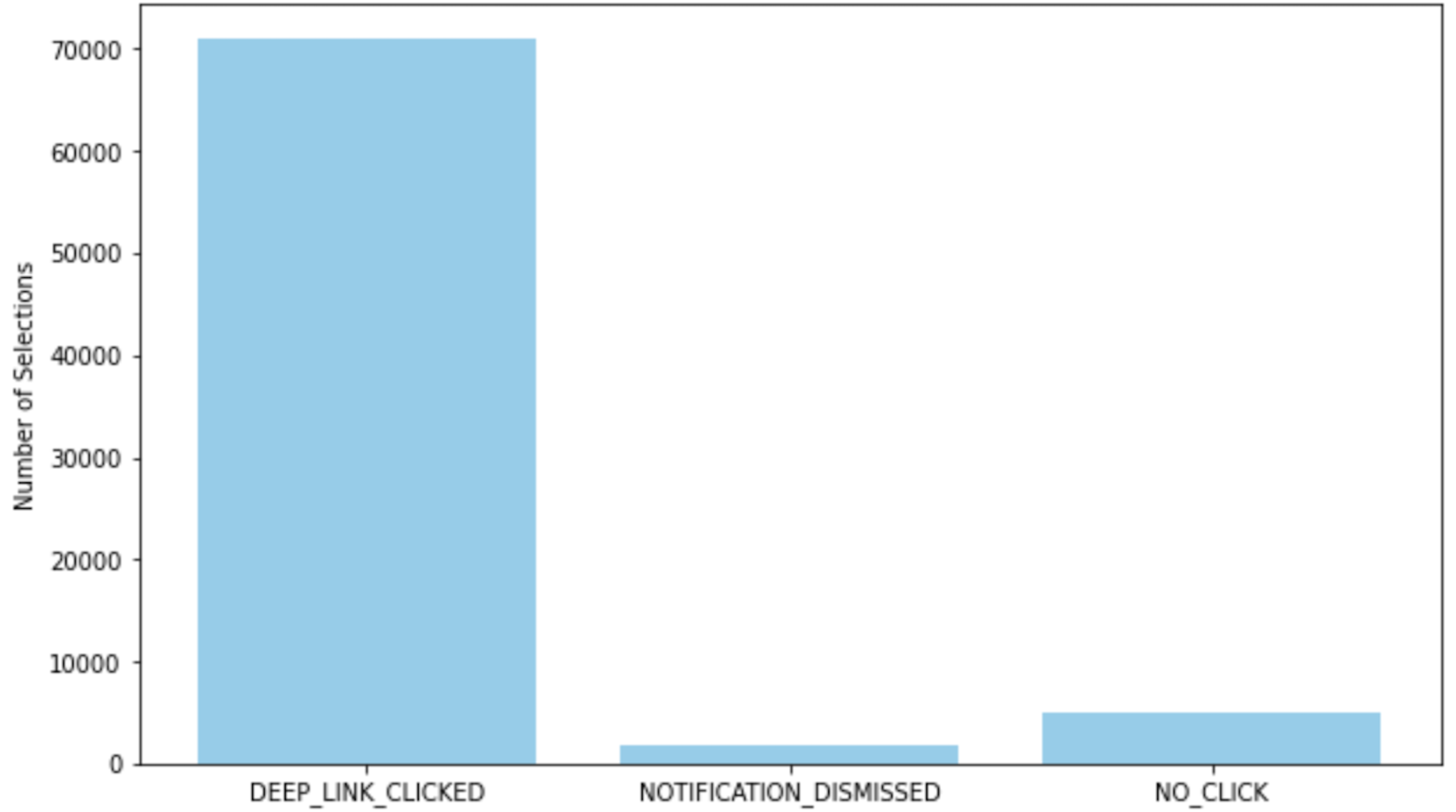}
        \captionsetup{width=0.45\textwidth}
        \caption{Feedback Distribution for Thompson Sampling}
        \label{fig:bandit_distribution}
    \end{figure}

\begin{figure}[h]
        \centering
        \includegraphics[width=0.5\textwidth]{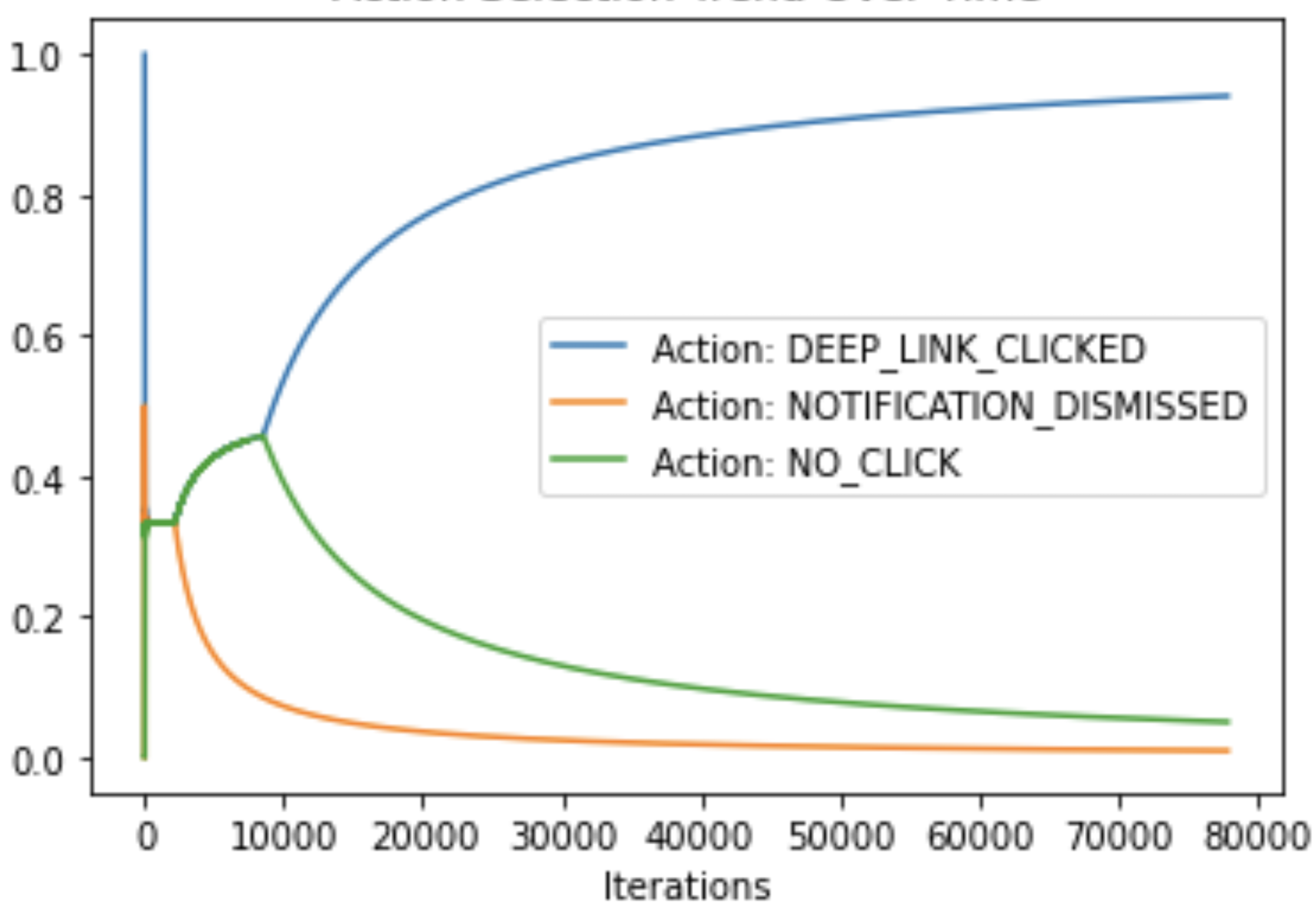}
        \captionsetup{width=0.45\textwidth}
        \caption{Feedback Trend Over Time for Thompson Sampling}
        \label{fig:Feedback_trend}
    \end{figure}

To evaluate the contextual bandit layer's performance, we conducted offline analysis. This analysis aimed to validate the bandit's ability to optimize recommendations based on contextual information and observed rewards. The key areas of analysis include:

\begin{itemize}
    \item \textbf{Feedback Distribution:}
    This analysis examines how frequently each feedback, i.e., action from sales representatives, (\texttt{DEEP\_LINK\_CLICKED}, \texttt{NOTIFICATION\_DISMISSED}, \texttt{NO\_CLICK}) was selected during the simulations. The distribution highlights the bandit's prioritization of high-reward feedback (Figure ~\ref{fig:bandit_distribution}).

    \item \textbf{Feedback Trends Over Time:}
    Temporal trends in Feedback percentages provide insights into the bandit's progression from exploration to exploitation, i.e., focusing on actions that leads high-reward feedback (Figure ~\ref{fig:Feedback_trend}).
\end{itemize}

\begin{figure}[h]
    \centering
    \includegraphics[width=0.5\textwidth]{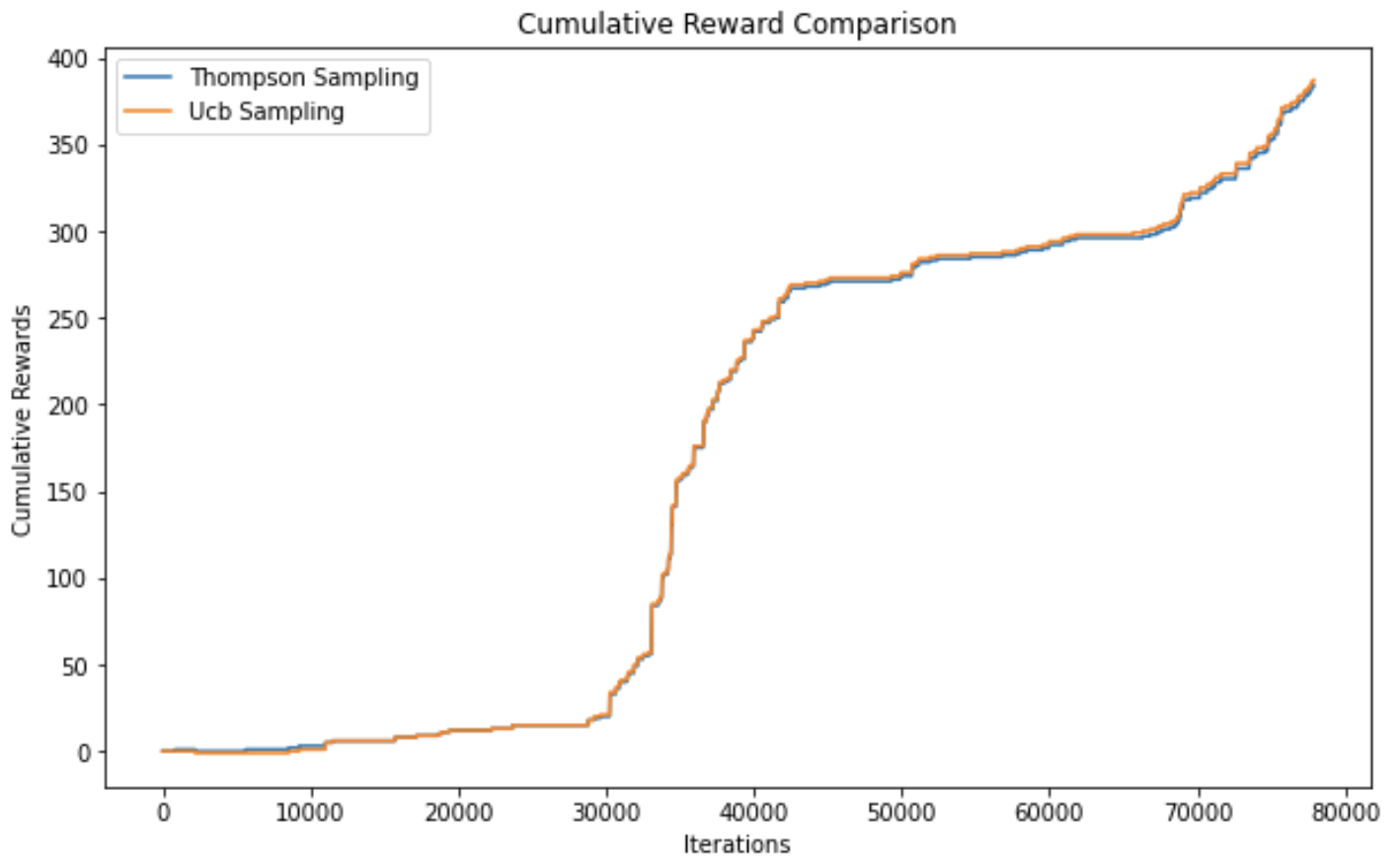}
    \caption{Comparison of Cumulative Rewards Between Thompson Sampling and UCB.}
    \label{fig:bandit_cumulative_rewards}
\end{figure}

For the bandit optimization, both Neural Thompson Sampling and Neural UCB approaches were evaluated to compare their effectiveness in terms of policy convergence, cumulative rewards, and their ability to balance exploration and exploitation. The cumulative rewards for both strategies are comparable over multiple iterations (Figure~\ref{fig:bandit_cumulative_rewards}). The current system adopts Thompson Sampling due to its low computational overhead per iteration leading to low latency while serving.

\section{Additional Details for Online Experiment}
\label{apx:online}
\subsection{Causal Inference Analysis and Assumption Validation}
\label{apx:did}

To validate the parallel trends assumption, we conducted a time-based A/A test (also known as pre-testing). In this approach, we used the same treatment and control groups as in our main analysis but ran placebo Difference-in-Differences tests at various points in the pre-treatment period. For each test, we treated an earlier time point as if it were the treatment period and evaluated whether the estimated treatment effect was close to zero. Since no intervention had actually occurred during these pre-periods, finding no significant differences supports the assumption that the treatment and control groups were following similar trends before the intervention. We also conditioned on important covariates like account size, engagement level, and past sales activity to ensure balance. 

\subsection{Survey on Explanation Layer}
\label{sec:explanatin_result}
We conducted a survey on the explainability of our recommendations across \textbf{142+ accounts}, achieving an overall \textbf{86\% satisfaction rate}, with strong positives on comprehensiveness and clarity.  

\noindent Key feedback highlights:  
\begin{itemize}
    \item Clear and actionable insights build trust in recommendations and uncover new opportunities.
    \item Consolidated insights save time, reducing the need to gather data from multiple sources.
    \item Explanations provide directional guidance, helping reps take right next steps in customer engagement.
    \item Sales reps' top signals for consideration align with the top important features from the explanation layer, reinforcing trust in the recommendation logic.
\end{itemize}

\end{document}